\documentclass[preprint,12pt]{elsarticle}
\usepackage{amsmath}
\usepackage{setspace}
\usepackage{makecell}
\usepackage{multibib}
\usepackage{graphicx,xcolor,comment}
\usepackage{pgfplots}
\pgfplotsset{compat=1.7}
\usepackage{dingbat}
\usepackage{multirow}
\usepackage{natbib}
\usepackage{lscape}
\usepackage{fancyhdr}
\usepackage{multicol}
\usepackage{adjustbox}
\usepackage{geometry}
\usepackage{caption}
\usepackage{rotating} 
\usepackage{float}
\usepackage{subcaption}
\usepackage{booktabs}
\usepackage{csquotes}
\usepackage{subcaption}
\usepackage{graphicx}  
\usepackage{listingsutf8}
\usepackage{hyperref}
\usepackage{cleveref}

\usepackage{amssymb}
\usepackage{amsmath}

\begin{document}

\begin{frontmatter}

\title{An Innovative Next Activity Prediction Using Process Entropy and Dynamic Attribute-Wise-Transformer in Predictive Business Process Monitoring}

\author[label1]{Hadi Zare} 
\author[label2]{Mostafa Abbasi}
\author[label1]{Maryam Ahang}
\author[label1,label2]{Homayoun Najjaran\corref{cor1}}

\cortext[cor1]{Corresponding author:}

\ead{najjaran@uvic.ca} 

\affiliation[label1]{organization={Department of Electrical and Computer Engineering, Faculty of Engineering and Computer Science, University of Victoria},
            addressline={3800 Finnerty Rd}, 
            city={Victoria},
            postcode={V8P 5C2}, 
            state={BC},
            country={Canada}}
\affiliation[label2]{organization={Department of Mechanical Engineering, Faculty of Engineering and Computer Science, University of Victoria},
            addressline={3800 Finnerty Rd}, 
            city={Victoria},
            postcode={V8P 5C2}, 
            state={BC},
            country={Canada}}


\begin{abstract}
Next activity prediction in predictive business process monitoring is crucial for operational efficiency and informed decision-making. While machine learning and Artificial Intelligence have achieved promising results, challenges remain in balancing interpretability and accuracy, particularly due to the complexity and evolving nature of event logs. This paper presents two contributions: (i) an entropy-based model selection framework that quantifies dataset complexity to recommend suitable algorithms, and (ii) the DAW-Transformer (Dynamic Attribute-Wise Transformer), which integrates multi-head attention with a dynamic windowing mechanism to capture long-range dependencies across all attributes. Experiments on six public event logs show that the DAW-Transformer achieves superior performance on high-entropy datasets (e.g., Sepsis, Filtered Hospital Logs), whereas interpretable methods like Decision Trees perform competitively on low-entropy datasets (e.g., BPIC\_2020\_Prepaid Travel Costs). These results highlight the importance of aligning model choice with dataset entropy to balance accuracy and interpretability.
\end{abstract}

\begin{highlights}
\item Entropy-driven model selection for optimal model selection balances accuracy and interpretability.
\item Novel DAW-Transformer model improves next activity prediction, especially on high-entropy event logs.
\item Dynamic windowing mechanism enhances long-range dependency capture in event logs.

\end{highlights}

\end{frontmatter}

\section{Introduction}

Process mining is a business process management (BPM) technique that provides insights extracted from event logs to improve organizational operations and service performance~\citep{burattin2015process, turner2012process, revina2023approach}. 
Among its applications, next activity prediction plays a crucial role by forecasting the continuation of ongoing or incomplete cases~\citep{dentamaro2023next, polato2018time}. Accurate forecasts enable proactive resource allocation, workflow optimization, and deviation prevention~\citep{sun2024next, pufahl2025resource, middelhuis2025learning, bukhsh2021processtransformer}. These predictive capabilities are particularly valuable in domains such as healthcare and manufacturing, where anticipating subsequent actions can significantly improve operational efficiency and outcomes. 

With the growing availability of event logs, organizations increasingly leverage predictive process monitoring to support decision-making, identify bottlenecks, and detect deviations from intended workflows~\citep{rivera2022multi, mehdiyev2020novel}. Machine learning (ML) and deep learning approaches, such as decision trees, random forests, long short-term memory (LSTM) networks, and convolutional neural networks (CNNs), have been explored for the next activity prediction. Among these, deep learning methods have gained significant attention due to their ability to capture complex patterns and dependencies~\citep{abbasi2025forlaps}.

Deep learning models, particularly Recurrent Neural Networks (RNNs) and Long Short-Term Memory (LSTM) networks, have been widely applied to sequence modeling tasks due to their ability to automatically learn hierarchical data representations~\citep{sun2024next,lecun2015deep}. In the context of process mining, these models leverage event logs to capture dependencies between activities and predict subsequent process steps~\citep{wang2023mitfm}. Despite their success, these models exhibit well-known limitations. Their performance degrades on long sequences, training and inference are computationally expensive due to their sequential structure, and they often rely primarily on activity sequences while neglecting additional attributes such as resources, roles, or case-specific variables~\citep{bukhsh2021processtransformer, musa2023prediction, di2019activity}.

Transformer architectures~\citep{vaswani2017attention} have emerged as a strong alternative by replacing recurrence with a self-attention mechanism capable of modeling long-range dependencies in parallel. Initially developed for natural language processing, Transformers have achieved state-of-the-art results across a wide range of AI domains and have recently been introduced to process mining~\citep{vaswani2017attention, wang2023mitfm, bukhsh2021processtransformer}. Their capacity to capture complex, multi-view relationships among event attributes makes them a promising direction for predictive process monitoring.

Additionally, there is a variety of models for predicting the next activity; however, selecting a suitable model based on dataset characteristics remains a challenge. Accuracy and interpretability are two crucial parameters when choosing a model. Here, interpretability refers to how easily the internal workings of a model can be understood, which is vital for explaining decisions to stakeholders.

Although different studies address some of these limitations, several important challenges remain:

\begin{itemize}

    \item \textbf{Limited use of additional attributes}: Modern event logs often contain attributes beyond the basic Case ID, Timestamp, and Activity. Many models fail to exploit these additional features, potentially losing important contextual information.

    \item \textbf{Fixed context windows}: Most existing methods use fixed window sizes, which may omit relevant information.

    \item \textbf{Lack of systematic model selection}: Choosing the best model requires balancing accuracy, efficiency, and interpretability. Current approaches rely on trial-and-error, which is costly and suboptimal~\citep{kumar2024transparency}.

\end{itemize}

To address these challenges, this paper makes two contributions:

\begin{itemize}
    \item \textbf{DAW-Transformer}: A multi-head Transformer model that integrates multiple event attributes throughout the prediction process and employs a dynamic windowing mechanism to handle long sequences efficiently.

    \item \textbf{Entropy-Driven Model Selection}: A novel approach that quantifies process flexibility and rigidness to recommend suitable models. High-entropy datasets benefit from DAW-Transformer, while low-entropy datasets favor interpretable models such as Decision Trees—balancing accuracy with explainability.
    
\end{itemize}

By addressing these challenges, our study improves predictive process monitoring and supports informed model selection.

The rest of this paper is organized as follows: Section 2 reviews related work on the next activity prediction. Section 3 provides preliminaries for understanding the approaches. Section 4 introduces the proposed methods and methodology, and Section 5 discusses experiments. Lastly, sections 6 and 7 cover the results, discussion, and conclusions.

\section{Related Work}

Over the past decade, predicting the next activity in BPM has attracted attention for improving organizational efficiency and supporting better decision-making. Numerous studies have focused on predicting the next activity in ongoing cases~\citep{sun2024next, impedovo2023next, weinzierl2020prescriptive, alaghbari2022activities, ceci2014completion}. Early approaches relied on traditional machine learning techniques. Models like Decision Trees were applied first due to their simplicity and interpretability~\citep{breiman2001random, song2015decision}. 
Decision trees have been widely adopted in predictive process monitoring scenarios, being used for tasks such as predicting Service Level Agreement (SLA) violations, Linear Temporal Logic (LTL) constraint violations, and process risks~\citep{conforti2015recommendation, di2018predictive, leitner2013data, tama2019empirical}. They have often been combined with other techniques like clustering~\citep{di2018predictive} or neural networks~\citep{leitner2013data} to improve accuracy. Similarly, Support Vector Machines (SVM) have also been popular for predicting process performance, risk, and unnecessary checks~\citep{kang2012periodic, verenich2016minimizing}, while evolutionary computing methods~\citep{marquez2017run} have shown potential for outperforming other ML algorithms in certain runtime prediction tasks. 

However, as event logs grew more complex, researchers increasingly turned to deep learning techniques to address these complexities~\citep{musa2023prediction, bukhsh2021processtransformer, abb2023discussion, di2019activity}. As the field has evolved, research in predictive process monitoring has focused not only on developing novel deep learning models for next-activity prediction but also on identifying effective methods for selecting the best model for each dataset. The next subsections first review the progress of deep learning in BPM and then discuss model selection methods, pointing out the gaps that inspire our work.


Initially, deep learning in BPM relied on RNNs for next-activity forecasting~\citep{abbasi2024review}. Although RNNs showed promising performance, they failed to remember the earlier context in lengthy sequences, and thus their performance on sequence prediction tasks was limited~\citep{wang2023mitfm}. To overcome this, LSTMs were applied, demonstrating improvements in sequence modeling~\citep{weinzierl2020xnap, krishna2018lstm, irbaz2024predicting, tax2017predictive}. However, LSTMs still faced challenges with long-range dependencies. However, LSTMs continued to face difficulties with long-range dependencies and are computationally expensive, often requiring significant time to capture input relationships~\citep{bukhsh2021processtransformer}. These limitations prompted researchers to explore other deep learning architectures.

In response to these challenges, convolutional neural networks (CNNs) were explored for next-activity prediction~\citep{di2019activity, rama2021deep}. This CNN-based approach outperformed LSTMs in both accuracy and computational efficiency. As a result, CNNs emerged as a strong alternative for modeling sequential data. Nevertheless, LSTM-based methods remained popular, with models such as Data-Aware Explainable Next Activity Prediction (DENAP)~\citep{aversano2023data}, which combines LSTMs with Layer-Wise Relevance Propagation (LRP), achieving high accuracy while enhancing interpretability.

Building on these developments, Transformers have emerged as a powerful solution to the limitations of RNNs and LSTMs by leveraging self-attention to model long-range dependencies. Introduced by~\citep{vaswani2017attention}, for neural machine translation, the Transformer architecture replaces the sequential recurrence of earlier models with a parallel attention-based design, enabling more efficient and scalable sequence processing. This shift enables the modeling of long-range dependencies and the extraction of generalized relationships between inputs and outputs. In self-attention, the importance of each token in a sequence is computed relative to a reference token, allowing the model to prioritize the most relevant elements when forming representations. For example, given a sequence of events, each is encoded as a query, key, and value vectors; the attention representation for a specific event is obtained by computing dot products between its query and all keys, producing attention weights that are applied to the corresponding values to form a weighted sum. Because each token’s attention is calculated independently, all positions can be processed in parallel, removing the need for recurrence~\citep{bukhsh2021processtransformer}.

To capture multiple types of relationships, such as semantic or temporal dependencies, the self-attention mechanism is executed in parallel across several projections, resulting in multi-head self-attention. Although initially developed for natural language processing, the Transformer is a domain-agnostic architecture that has proven effective for diverse sequence modeling tasks, including predictive business process monitoring. Its ability to capture complex, multi-view dependencies makes it a strong candidate for advancing next-activity prediction in BPM. Models such as the Multi-View Information Fusion Method (MiFTMA) and the Multi-Task Learning Guided Transformer Network (MTLFormer) have shown notable improvements in capturing long-term dependencies, reducing complexity, and improving prediction accuracy~\citep{wang2023mitfm, wang2023mtlformer}. However, these Transformer-based approaches rely on a sliding window mechanism, which restricts their ability to capture broader process behaviors because they only consider fixed-length segments of event traces. Similarly, the ProcessTransformer~\citep{bukhsh2021processtransformer} demonstrated the potential of applying Transformer architectures to predictive business process monitoring, it primarily focused on modeling activity sequences using fixed-size context windows and did not fully exploit all available event attributes, such as resources or case-level features.

While traditional machine learning methods, such as decision trees, offered simplicity and interpretability, they struggled with complex event logs and failed to capture temporal dependencies. Deep learning methods addressed some of these limitations, yet RNNs and LSTMs suffered from vanishing gradients, long-range dependency issues, and high computational costs, while CNNs were limited to modeling local patterns. Transformer-based models significantly advanced next-activity prediction by leveraging self-attention for long-range dependencies, but they often rely on fixed sliding windows and focus primarily on activity sequences, overlooking valuable contextual attributes such as resources, time, and case-level features. To address these limitations, our work introduces a dynamic, context-aware approach, DAW-Transformer, that moves beyond fixed-window representations and integrates multiple event attributes to more effectively capture complex process behaviors for next-activity prediction.

Building on this, a further challenge lies in determining when such complex models are essential, as not all event logs exhibit the same degree of variability~\citep{raschka2018model}. Prior research highlights the importance of aligning model complexity with dataset variability to balance accuracy, efficiency, and interpretability~\citep{domingos2012few, kim2024explaining, drodt2023predictive}. In particular, measures of entropy have long been used to quantify uncertainty in information systems~\citep{jung2008measuring, jung2011entropy} and have been applied in domains such as computer vision, natural language processing~\cite{rahane2020measures} to characterize data complexity and guide model selection.


Event logs in process mining differ significantly in their structure: some exhibit high variability and complexity (high entropy), whereas others follow more stable and predictable patterns (low entropy)~\cite{revina2023approach}. Despite this diversity, existing studies rarely apply a systematic approach to model selection, often defaulting to deep learning methods regardless of dataset complexity. This practice introduces unnecessary computational overhead and reduces interpretability.

To address this gap, we propose an \textbf{Entropy-Driven Model Selection} strategy that quantifies dataset complexity and ensures optimal model selection. Unlike previous works that rely on trial and error~\citep{raschka2018model}, our approach uses normalized Shannon entropy. The framework recommends complex models such as the DAW-Transformer for high-entropy datasets, while suggesting simpler, interpretable models such as Decision Trees for low-entropy datasets. This approach ensures deep learning is applied only when necessary, providing a systematic balance between accuracy, interpretability, and efficiency.

\section{Preliminaries}
This section introduces foundational concepts relevant to process entropy and process mining for a better understanding of our proposed method.
\subsection{Event log}
Event logs record data about various event types and their timestamps, typically collected during the operation of modern industrial systems and machines. These logs are crucial for process analysis, enabling predictive insights, optimization, and proactive responses that improve system efficiency and reliability~\citep{huang2021deep}. Table~\ref{tab:event_log} presents a sample event log from the BPIC\_2012\_A event logs, highlighting the fundamental details of an event, including: \textbf{CaseID} (identifying a process instance), \textbf{Activities} (representing process steps), and \textbf{Timestamps} (marking when events occur).

\begin{table}[ht]
    \centering
    \caption{A sample event log of the BPIC\_2012\_A event logs.}
    \resizebox{\textwidth}{!}{ 
    \begin{tabular}{@{}ccccccc@{}} 
        \toprule
        \textbf{Case ID} & \textbf{Activity} & \textbf{Timestamp} & \textbf{Lifecycle} & \textbf{Amount Requested} & \textbf{org:resource} & \textbf{REG\_DATE} \\ 
        \midrule
        173688 & A\_SUBMITTED & 2011-10-01 00:38:44 & COMPLETE & 20,000 & 112 & 2011-10-01 \\
        173688 & A\_PARTLYSUBMITTED & 2011-10-01 00:38:44 & COMPLETE & 20,000 & 112 & 2011-10-01 \\
        173688 & A\_PREACCEPTED & 2011-10-01 00:39:37 & COMPLETE & 20,000 & 112 & 2011-10-01 \\
        173688 & A\_ACCEPTED & 2011-10-01 11:42:43 & COMPLETE & 20,000 & 10862 & 2011-10-01 \\
        173688 & A\_FINALIZED & 2011-10-01 11:45:09 & COMPLETE & 20,000 & 10862 & 2011-10-01 \\
        173688 & A\_REGISTERED & 2011-10-13 10:37:29 & COMPLETE & 20,000 & 10629 & 2011-10-01 \\
        173688 & A\_APPROVED & 2011-10-13 10:37:29 & COMPLETE & 20,000 & 10629 & 2011-10-01 \\
        173688 & A\_ACTIVATED & 2011-10-13 10:37:29 & COMPLETE & 20,000 & 10629 & 2011-10-01 \\
        \bottomrule
    \end{tabular}
    \label{tab:event_log}
    }
\end{table}

\subsection{Process Entropy}
\label{Entropy}
Process entropy quantifies the uncertainty in business process execution~\citep{jung2008measuring}. High process entropy indicates unpredictability, making scheduling and resource allocation more challenging. In information theory, uncertainty is typically quantified using information entropy, commonly known as Shannon’s entropy ~\citep{jung2008measuring}. It is defined as:
\begin{equation}
H(X) = - \sum_{i=1}^n p(x_i) \log(p(x_i))
\end{equation}
In this expression, \( X \) represents a discrete random variable that can assume possible values \( x_1, x_2, \ldots, x_n \) with corresponding probabilities \( p(x_1), p(x_2), \ldots, p(x_n) \). For \( 1 \leq i \leq n \), the probabilities satisfy \( p(x_i) \geq 0 \) and $\sum_{i=1}^n p(x_i) = 1$.
In process mining, we are particularly interested in conditional entropy, which measures the uncertainty of the next activity given the current one. It is defined as:
\begin{equation}
H(Y \mid X) = - \sum_{x \in X} p(x) \sum_{y \in Y} p(y \mid x) \log_2 p(y \mid x)
\end{equation}
Conditional entropy captures the predictability of a process: lower values indicate that the process follows a more deterministic path, while higher values suggest greater flexibility. In this study, we compute conditional entropy based on transition frequencies extracted from the event log, as detailed in the Methodology section.

\subsection{Trace}

In event logs, it is possible to identify multiple cases. A case is characterized as a sequence of events, commonly termed a ``trace" in the literature. Let T denote a trace, represented as:

\begin{equation}
T = \langle e_1, e_2, e_3, \ldots, e_n \rangle
\end{equation}

where each \( e_i \) signifies a recorded event. Each event \( e_i \) is linked to multiple attributes, the quantity of which may differ based on the particular event log under examination~\citep{bolt2017finding}.

\section{Methodology}
This section presents the details of the proposed DAW-Transformer and the entropy-driven model selection method, both designed to predict the next activity efficiently while balancing interpretability, accuracy, and considering all attributes at any given time.

\subsection{Multi-Transformer}
\label{sec:daw_transformer_methodology}

The Multi-Transformer is designed to capture complex dependencies in sequential data while integrating multiple feature types. By leveraging attention mechanisms, it can model relationships within sequences more effectively than traditional approaches.

\subsubsection{Multi-Feature Embedding and Position Encoding}
This component aims to comprehensively represent each sequence by embedding categorical and numerical features while incorporating positional encoding to capture the order of events. Embedding is crucial for the model to understand the relationships between categorical and numerical features and their temporal evolution. Positional encoding is significant as the attention mechanism lacks awareness of the sequence order. By assigning a specific position to each, the model can naturally understand the progression and order of sequence and, in turn, have a deeper grasp of temporal dependencies~\citep{vaswani2017attention}.

\subsubsection{Architecture}
\label{Transformer}
The transformer encoder block is central to this model, enabling the integration of numerical data for prediction. This block begins by applying multi-head self-attention, which captures the relationships within each sequence. The attention mechanism is computed as follows: 
\begin{equation}
Attention (Q, K, V) = \text{softmax}\left(\frac{Q K^T}{\sqrt{d_k}}\right) V 
\end{equation}

$V$, with $Q$, $K$, and $V$ representing queries, keys, and values, respectively, and $d_k$ the key dimension~\citep{vaswani2017attention}. A residual connection and layer normalization stabilize learning and enhance performance.

Next, a feed-forward network (FFN) introduces non-linearity and transforms the data using the following equation:
\begin{equation}
\text{FFN}(x) = \text{ReLU}(x W_1 + b_1) W_2 + b_2 
\end{equation}
The embeddings are then transformed, flattened, and concatenated with additional numerical features to enhance the input representation~\citep{vaswani2017attention}.

Finally, a dense output layer with a softmax activation function generates a probability distribution over the prediction classes. This approach effectively integrates sequential and numerical information, ensuring comprehensive and accurate predictions.


\subsection{DAW-Transformer}

To tackle the challenge of predicting the next activity in high-entropy, complex business processes, we propose the DAW-Transformer. This model builds upon the baseline Multi-Transformer architecture presented in Section~\ref{Transformer}. While the Multi-Transformer typically relies on a fixed-size sliding context window (e.g., a limited number of recent events), the DAW-Transformer dynamically determines the input window based on the characteristics of each dataset. Specifically, it identifies the maximum sequence length (i.e., the length of the longest case trace) and uses this as the window size for that dataset. All other case traces are padded accordingly. This approach allows the model to adjust to the complexity and structure of each dataset, ensuring that sufficient historical context is preserved without truncating valuable information. Additionally, the DAW-Transformer incorporates multi-modal features, such as categorical (activities, resources), temporal, and numerical (Age, CRP, LacticAcid, Leucocytes) in Sepsis event logs, which allows for richer contextual modeling. This comprehensive, full-history approach provides the model with a more holistic view of each case’s progression, enabling superior performance in high-entropy, complex business processes.

To clearly highlight the differences between the DAW-Transformer and the baseline vanilla Transformer, Table~\ref{tab:transformer_comparison} summarizes the key distinctions in sequence handling, attribute usage, and suitability for complex, high-entropy logs. While the vanilla Transformer relies on a fixed sliding window and limited event attributes, the DAW-Transformer processes full-length traces, pads shorter sequences, and incorporates all available event attributes, enabling improved predictive performance. This comparison emphasizes the novelty and advantages of our approach~\citep{vaswani2017attention}.

\begin{table}[ht]
    \centering
    \renewcommand{\arraystretch}{1.6} 
    \caption{Comparative overview of the Vanilla Transformer and DAW-Transformer architectures.}
    \label{tab:transformer_comparison}
    \resizebox{\textwidth}{!}{%
    \begin{tabular}{lcc}
        \toprule
        \textbf{Dimension} & \textbf{Vanilla Transformer} & \textbf{DAW-Transformer} \\
        \midrule
        Sequence Handling & Fixed sliding window & \makecell{Full-length sequence (longest trace);\\ shorter traces padded}   \\
        Event Attributes & Limited (e.g., activity only) & \makecell{Comprehensive: categorical, numerical,\\ and temporal attributes}  \\
        Contextual Scope & Limited to recent events & Entire trace context \\
                \bottomrule
    \end{tabular}}
\end{table}


\subsubsection{Data Preparation}

To illustrate the preprocessing procedure, we refer to the \textit{Sepsis} event logs, which combine control-flow information with clinical data. Categorical variables, including activities and resources, are first label-encoded and subsequently transformed into dense representations through embedding. Temporal features are derived from event timestamps, expressed as fractional hours relative to the case start, and normalized to the $[0,1]$ range using min–max scaling. Clinical attributes such as \textit{Age}, \textit{CRP}, \textit{LacticAcid}, and \textit{Leucocytes} are z-score normalized, while missing values are imputed using a combination of forward- and backward-filling within each case.  

\subsubsection{Sequence Preparation and Padding}

The data are first grouped by case, yielding sequences of tuples in which each event is represented together with its corresponding feature values. From these sequences, input–output pairs are constructed by generating sub-sequences for each case: the inputs comprise all features up to the current event, while the outputs correspond to the subsequent activity. To facilitate learning across cases, separate lists are maintained for each input feature as well as for the output. In order to ensure consistent input dimensions for the model, all sequences are subsequently padded to a uniform length.


\subsubsection{Model Architecture}

The DAW-Transformer architecture integrates categorical, temporal, and numerical inputs into a unified predictive framework. Categorical features (activities and resources) are mapped into dense vectors through embedding layers, with trainable positional embeddings added to capture event order. Sequence dependencies are modeled using stacked Transformer encoder blocks, which include multi-head self-attention, feed-forward layers, residual connections, and normalization. Numerical features, comprising both time-based and clinical attributes, bypass embedding and are instead processed directly through dense layers. These numerical representations are reshaped and concatenated with the output of the Transformer blocks. Finally, all feature representations are fused and passed through fully connected layers with dropout for regularization, culminating in a softmax classifier for next-activity prediction.

\subsubsection{Training and Evaluation}

The dataset was split into training (64\%), validation (16\%), and testing (20\%) sets. The validation set was used for early stopping and model checkpointing, while the test set was reserved for the final unbiased evaluation. Model performance was evaluated using Accuracy, Precision, Recall, and F1-score. A classification report and confusion matrix were generated to interpret the model’s behavior across different activities.

\subsection{Entropy Computation}
\label{sec:Entropy Computation}
In this section, we describe the approach used to compute the conditional entropy of activity transitions within the business process. This metric helps quantify the uncertainty in predicting the next activity given the current one, providing insight into the level of determinism in the process behavior.

\subsubsection{Data Preparation}
To avoid data leakage, the entropy used for model selection is calculated only on the training and validation data, which together constitute 80\% of the dataset. By excluding the test set, we ensure that information from unseen cases does not influence model selection. This allows the entropy to reflect the inherent uncertainty of the process based solely on the data used for training, while keeping the test set completely independent for unbiased evaluation. 

Each event log consists of several attributes such as activity, timestamp, case ID, and context-specific features (\textit{e.g.,} resource). During preprocessing, categorical attributes are encoded, and all data is standardized to ensure compatibility with machine learning algorithms. 

 Event logs stored in a CSV file are used, where each row represents an event and contains at least the following columns:

\begin{itemize}
    \item \texttt{case:concept:name}: the unique identifier for each process instance (case)
    \item \texttt{concept:name}: the activity name
\end{itemize}

The event logs are loaded using the pandas library and group activities by their corresponding case ID to reconstruct \textit{traces}, i.e., sequences of activities that occurred within each process instance.

\subsubsection{Transition Extraction}

To capture the sequential relationships between activities, all \textit{bigrams} are extracted, i.e., ordered pairs of consecutive activities \((a, b)\), from each trace. These bigrams represent observed transitions between steps in the process.

\begin{verbatim}
bigrams = []
for trace in traces:
    bigrams.extend([(trace[i], trace[i+1]) for i in range(len(trace) - 1)])
\end{verbatim}

We count:
\begin{itemize}
    \item The frequency of each bigram (i.e., transition): \texttt{Counter((a, b))}
    \item The frequency of each activity as the current activity \(a\)
\end{itemize}


\subsubsection{Conditional Entropy Calculation}

The uncertainty of predicting the next activity given the current one can be formally expressed as the conditional entropy.

In our implementation, $p(x)$ corresponds to the empirical probability of observing activity $x$ as the current activity, and $p(y \mid x)$ denotes the probability of transitioning from $x$ to $y$, estimated as:

\begin{equation}
    p(y \mid x) = \frac{\text{count}(x, y)}{\text{count}(x)}.
\end{equation}

Thus, while the transition probabilities $p(y \mid x)$ are directly computed from the event log, the entropy formulation aggregates these probabilities into a single metric of conditional uncertainty, which we refer to as conditional entropy.

\subsubsection{Normalization}
\label{sec:Normalization}

To compare entropy values across datasets with varying numbers of unique activities, the conditional entropy is normalized using the base-2 logarithm of the total number of unique activities~\cite{wilcox1967indices}.  This normalization accounts for the fact that event logs may have high conditional entropy even with a small number of distinct activities, while other logs with the same entropy may involve many more activities, indicating that some activities occur rarely. Normalization ensures that entropy values from different event logs can be compared on the same scale, as they are adjusted based on the number of activities in each log. This ensures that the normalized entropy value lies within the interval $[0, 1]$, facilitating consistent interpretation regardless of the dataset size.

\subsubsection{Entropy-Driven Model Selection Approach}

Building on the entropy computation described in Section~\ref{sec:Entropy Computation}, we introduce an adaptive model selection strategy guided by the level of process entropy. The objective is to align model complexity with the behavioral variability observed in the dataset. Based on the normalized entropy values calculated earlier, we distinguish between structured and unstructured process behaviors (as interpreted in Section~\ref{sec:Normalization}). This distinction forms the basis of our model selection approach:

\begin{itemize}
    \item \textbf{Low Entropy} suggests structured and predictable process behavior. In such cases, simpler models like Decision Trees are typically sufficient to achieve high accuracy.
    \item \textbf{High Entropy} indicates unstructured, complex, and less predictable process behavior. For these datasets, more sophisticated models such as the DAW-Transformer are preferred, due to their ability to capture intricate sequential patterns.
\end{itemize}

To distinguish between structured and unstructured processes, entropy values are interpreted along a spectrum of predictability, with higher values indicating more unstructured processes.

\begin{figure*}[htbp]
    \centering
 
    \hspace*{0pt}
    \vspace{10pt}
    \includegraphics[width=0.9\textwidth]{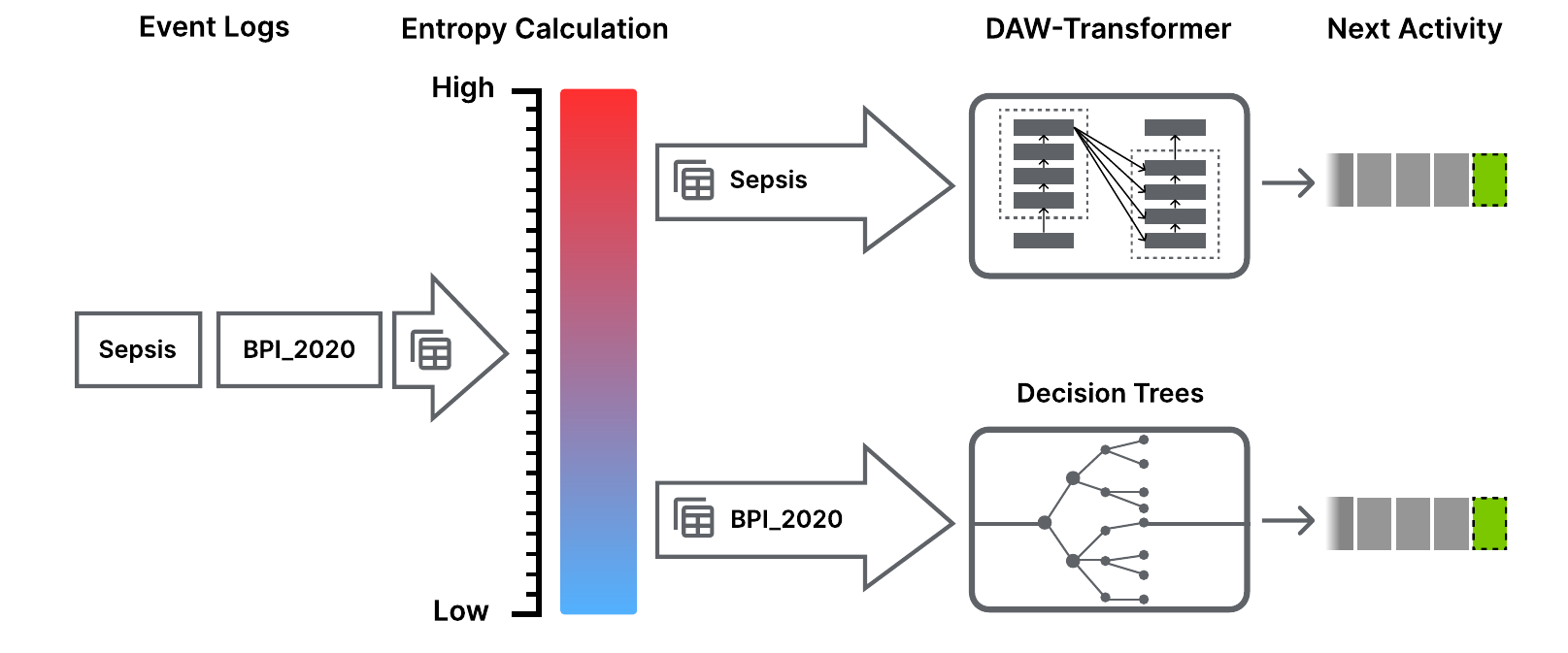}
    \caption{\centering
    Entropy-Driven Model Selection: Datasets with higher entropy benefit from DAW-Transformer, while those with lower entropy can rely on Decision Trees for interpretability with comparable performance.}
    \label{fig:framework}
\end{figure*}

This entropy-driven selection strategy enables an informed and adaptive modeling process, ensuring that the predictive model aligns with the underlying complexity of the data.~\Cref{fig:framework} illustrates the overall methodology, showing how entropy estimation supports model selection in next activity prediction tasks.

\vspace{10pt}

\section{Experiments}
This section presents an experimental evaluation of the DAW-Transformer model, comparing its performance with that of CNN-LSTM, CNN-BiLSTM, Multi-Transformer (limited window), XGBoost, Decision Trees, and Random Forest. Additionally, the datasets are analyzed based on their process entropy to determine the most suitable model for each dataset. 

\subsection{Datasets}
The experiments were carried out on seven publicly available datasets commonly used in process mining. Some relevant statistics from these logs are shown in Table~\ref{tab:dataset_properties}. These include the number of cases,  the number of events, the number of different activities, as well as the average and maximum case lengths. The datasets were chosen as the foundation for this study, with each representing a distinct real-world process. The following provides a brief description of each event log:

\textbf{Sepsis}: This real-world event log includes events of sepsis cases from a hospital, documented by the ERP (Enterprise Resource Planning) system. Each case in the log represents a patient's journey through the hospital~\cite{https://doi.org/10.4121/uuid:915d2bfb-7e84-49ad-a286-dc35f063a460}.

\textbf{Filtered Hospital logs}: Real-life event log from a Dutch academic hospital, originally published for the first Business Process Intelligence Contest (BPIC 2011)~\cite{vandongen2011hospital}. The original event log contains 624 different activities with a normalized entropy of 0.236. In this study, the log is filtered to include only activities that occurred more than 400 times. After filtering, the number of activities is reduced to 64, and the normalized entropy increases to 0.313, indicating high entropy and making it suitable for our work.

\textbf{NASA}: The event log is obtained by instrumenting the NASA Crew Exploration Vehicle (CEV) class using the XPort tool. It records method-call level events describing a single run of an exhaustive unit test suite for the CEV example, which is documented in the JPF Statechart project. The life-cycle information corresponds to method calls (start) and returns (complete), thus capturing a method-call hierarchy~\cite{leemans2017nasa}. 

\textbf{BPIC\_2012\_A}: The BPIC Challenge 2012\_A dataset is an event log recording a real-world loan application process~\cite{vandongen2012BPI2012}.

\textbf{Helpdesk}: This dataset comprises events from the ticket management process of the help desk of an Italian software company. Each case in the log begins with a new ticket entry into the ticket management system and concludes with the resolution of the issue and the closing of the ticket~\cite {https://doi.org/10.17632/39bp3vv62t.1}.

\textbf{BPIC\_2020\_Prepaid Travel Costs}: This file includes events associated with prepaid travel expenses for the parent item~\cite{https://doi.org/10.4121/uuid:5d2fe5e1-f91f-4a3b-ad9b-9e4126870165}.

\begin{table}[ht]
    \centering
    \caption{General properties of each dataset.}
    \label{tab:dataset_properties}
    \resizebox{\textwidth}{!}{%
    \begin{tabular}{lccccc}
        \toprule
        \textbf{Event log} & \textbf{\makecell{Num.\\ cases}} & \textbf{\makecell{Num.\\ events}} & \textbf{\makecell{Num.\\ activities}} & \textbf{\makecell{Avg. case\\ length}} &
        \textbf{\makecell{Max. case\\ length}}\\
        \midrule
        Sepsis & 1,049 & 15,214 & 16 & 14.48 & 185\\
        Filtered\_Hospital log & 1,142 & 136,023 & 64 & 119.11 & 1,611\\
        NASA & 2,566 & 73,638 & 47 & 28.70 & 50\\
        BPIC\_2012\_A & 13,087 & 60,849 & 10 & 4.65 & 8 \\
        Helpdesk & 4,580 & 21,348 & 14 & 4.66 & 15 \\
        BPIC\_2020\_Prepaid travel cost & 2,099 & 18,246 & 29 & 8.69 & 21\\
        
        \bottomrule
    \end{tabular}}
\end{table}

\subsection{Hyperparameter Setup}
\Cref{tab:hyperparameters} presents the hyperparameters used for the DAW-Transformer model on the Sepsis dataset. The model employs an embedding dimension of 256 to effectively capture feature representations, with 8 attention heads to enhance the learning of sequential dependencies. A feed-forward dimension of 256 ensures sufficient model capacity for processing complex patterns in activity sequences. The Adam optimizer is utilized for efficient gradient-based optimization, while a batch size of 16 is chosen to accommodate the variable-length sequences in the dataset. The model is trained for 50 epochs with a validation split of 0.2, ensuring a balanced evaluation of its generalization ability. These hyperparameter choices are designed to optimize performance while maintaining stability during training. Additionally, early stopping with a patience of 10 and model checkpointing based on validation loss were employed to prevent overfitting and retain the best model. Layer normalization, ReLU activation in the feed-forward network, and no dropout were applied within the Transformer blocks.

\begin{table}[ht]
    \centering
    \caption{Hyperparameters for DAW-Transformer Model on the Sepsis dataset.}
    \begin{tabular}{@{}ll@{}}
        \toprule
        \textbf{Hyperparameter}                    & \textbf{Value} \\ \midrule
        Embedding dimension (embed\_dim)           & 256            \\
        Number of Transformer heads (num\_heads)   & 8              \\
        Feed-forward dimension (ff\_dim)           & 256            \\
        Activation function                        & ReLU           \\
        Normalization                              & LayerNorm ($\epsilon=10^{-6}$) \\
        Positional encoding                        & Learned embeddings \\
        Optimizer                                  & Adam           \\
        Loss function                              & Sparse categorical cross-entropy \\
        Batch size                                 & 16             \\
        Number of epochs                           & 50 (with early stopping) \\
        Early stopping patience                    & 10             \\
        Validation split                           & 0.2            \\
        Model checkpointing                        & Best model (val\_loss) saved \\ \bottomrule
    \end{tabular}
    \label{tab:hyperparameters}
\end{table}

For the Sepsis dataset, a CNN-BiLSTM model was used, with carefully tuned hyperparameters to enhance performance, as shown in~\Cref{tab:model_hyperparameters}. Key parameters included an initial filter size of 64 for the first convolutional layer, which progressively increased to 256 in subsequent layers, facilitating the extraction of complex features. To counteract overfitting, dropout values of 0.4 and 0.5 were added. To extract temporal relations in the sequence, a 128-unit bidirectional LSTM layer was added. The Adam optimizer with a learning rate of 0.001 was employed, and the model was trained over 300 epochs with a batch size of 32; early stopping and a learning rate scheduler were used to refine the training process and enhance generalization.

\begin{table}[htbp]
    \centering
    \caption{Hyperparameters for CNN-LSTM model on sepsis dataset.}
    \resizebox{\textwidth}{!}{%
    \begin{tabular}{@{}l|l|l|l|l|l@{}}
        \toprule
        \textbf{Hyperparameter}        & \textbf{Value} & \textbf{Hyperparameter}        & \textbf{Value} & \textbf{Hyperparameter}        & \textbf{Value} \\ \midrule
        Filters in 1st Conv Layer      & 64             & Optimizer                      & Adam           & Dropout rate after 1st Conv    & 0.4            \\
        Kernel size in 1st Conv Layer  & 3              & Learning rate                  & 0.001          & Filters in 2nd Conv Layer      & 128            \\
        Pool size in 1st Max Pooling   & 1              & Batch size                     & 32             & Kernel size in 2nd Conv Layer  & 3              \\
        Dropout rate after 1st Conv    & 0.4            & Number of epochs               & 300            & Pool size in 2nd Max Pooling   & 1              \\
        Filters in 2nd Conv Layer      & 128            & Validation split               & 0.2            & Dropout rate after 2nd Conv    & 0.5            \\
        Kernel size in 2nd Conv Layer  & 3              & Early stopping patience        & 30             & Filters in 3rd Conv Layer      & 256            \\
        Pool size in 2nd Max Pooling   & 1              & Learning rate scheduler patience & 10            & Kernel size in 3rd Conv Layer  & 3              \\
        Dropout rate after 2nd Conv    & 0.5            & Learning rate scheduler factor & 0.5            & Pool size in 3rd Max Pooling   & 1              \\
        Filters in 3rd Conv Layer      & 256            & Learning rate scheduler min lr & 1e-6           & Dropout rate after 3rd Conv    & 0.5            \\
        Units in LSTM Layer            & 128            & Output Layer activation        & Softmax        & Units in Dense Layer           & 100            \\
        L2 Regularization in Dense     & 0.02           & Dropout rate after Dense Layer & 0.6            &                                &                \\ \bottomrule
    \end{tabular}%
    }
    \label{tab:model_hyperparameters}
\end{table}

\subsection{Evaluation Metrics}

To assess the performance of the proposed model in predicting the next activity, several widely used evaluation metrics are employed. These metrics provide complementary perspectives on classification quality, especially in datasets with imbalanced classes. Specifically, we consider \textbf{accuracy}, \textbf{precision}, \textbf{recall}, and \textbf{F1-score}.

\subsubsection{Accuracy}
Accuracy measures the overall proportion of correctly predicted events across all classes~\citep{evermann2016deep, khan2021deepprocess}. Let \( \mathit{tp}_i \), \( \mathit{fp}_i \), \( \mathit{tn}_i \), and \( \mathit{fn}_i \) denote the true positives, false positives, true negatives, and false negatives for class \( i \), respectively. Let \( s_i \) be the number of events in class \( i \) and \( n = \sum_{i=1}^{l} s_i \) the total number of events in the dataset. The weighted average accuracy is computed as:

\begin{equation}
\text{Accuracy} = \frac{1}{n} \sum_{i=1}^{l} s_i \frac{tp_i + tn_i}{tp_i + tn_i + fp_i + fn_i}
\end{equation}

\subsubsection{Precision}
Precision quantifies the proportion of correctly predicted events for a given class among all events predicted as belonging to that class~\citep{foody2023challenges}. It is defined for class \( i \) as:

\begin{equation}
\text{Precision}_i = \frac{tp_i}{tp_i + fp_i}
\end{equation}

The overall precision can be reported as a weighted average across all classes:

\begin{equation}
\text{Precision} = \frac{1}{n} \sum_{i=1}^{l} s_i \cdot \text{Precision}_i
\end{equation}

\subsubsection{Recall}
Recall measures the proportion of correctly predicted events for a class among all actual events of that class~\citep{foody2023challenges}. For class \( i \), it is defined as:

\begin{equation}
\text{Recall}_i = \frac{tp_i}{tp_i + fn_i}
\end{equation}

Similarly, the weighted average recall across all classes is:

\begin{equation}
\text{Recall} = \frac{1}{n} \sum_{i=1}^{l} s_i \cdot \text{Recall}_i
\end{equation}

\subsubsection{F1-score}
The F1-score is the harmonic mean of precision and recall, providing a balance between the two metrics~\citep{bukhsh2021processtransformer}. For class \( i \), it is computed as:

\begin{equation}
\text{F1}_i = 2 \cdot \frac{\text{Precision}_i \cdot \text{Recall}_i}{\text{Precision}_i + \text{Recall}_i}
\end{equation}

The weighted F1-score across all classes is:

\begin{equation}
\text{F1-score} = \frac{1}{n} \sum_{i=1}^{l} s_i \cdot \text{F1}_i
\end{equation}

These four metrics together provide a comprehensive evaluation of the model’s predictive performance, accounting for both overall correctness and the ability to identify individual classes correctly.

\section{Results}

The resulting values of conditional entropy provided a numerical indicator of process predictability. A low entropy value reflects structured behavior where the current activity largely determines the next activity, while high entropy values indicate unstructured or flexible processes with less predictable activity sequences.

\begin{figure*}[htbp]
    \centering
 
    \hspace*{0pt}
    \vspace{10pt}
    \includegraphics[width=1\textwidth, page = 2]{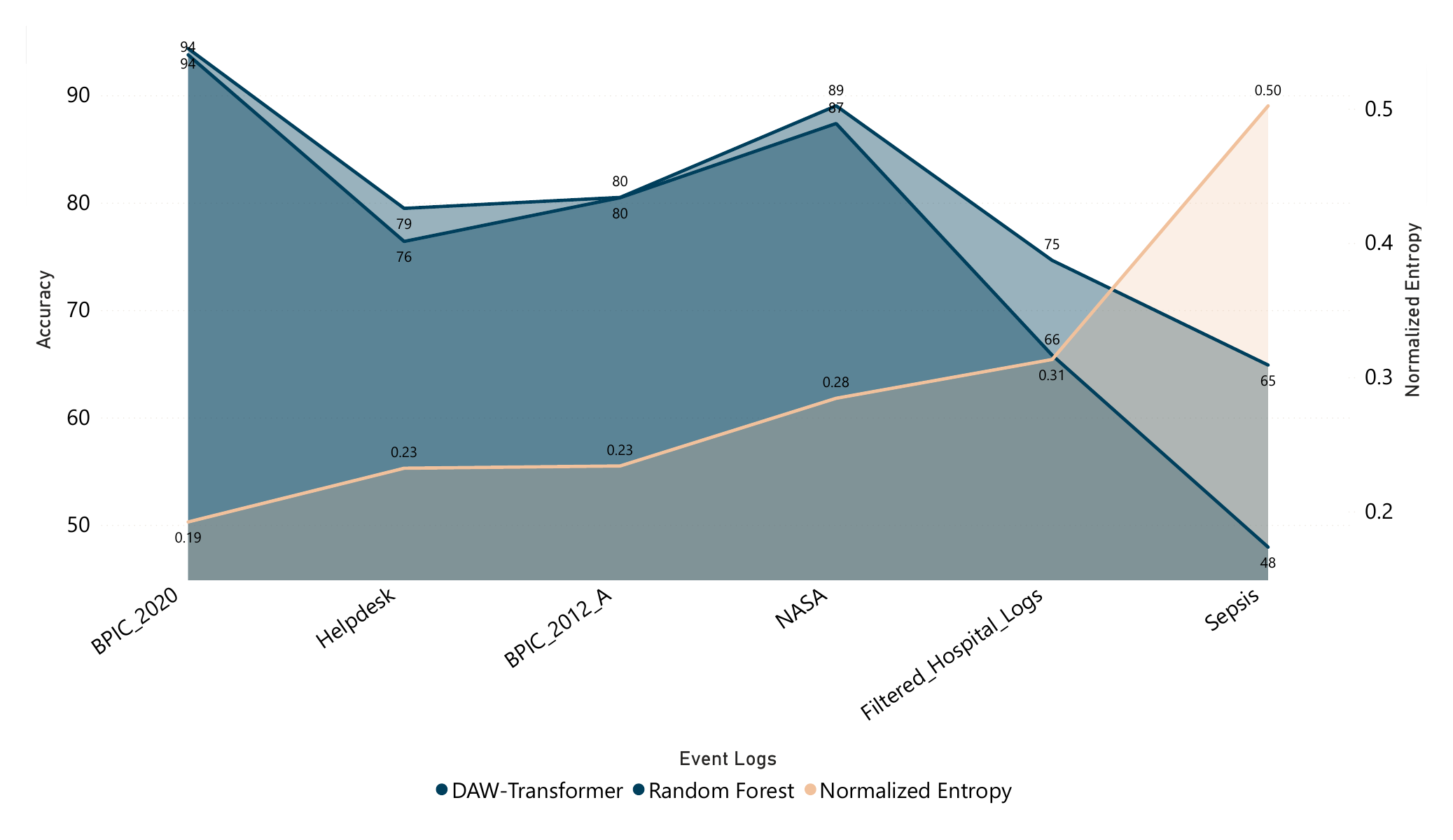}
    \caption{\centering
         Model accuracy across event logs: Random Forest, and DAW-Transformer perform similarly in low-entropy logs, while DAW-Transformer shows clear advantages in high-entropy logs.}
    \label{fig:results:charts2}
\end{figure*}

Our experimental observations (Figure~\ref{fig:results:charts2}) revealed a consistent pattern which in datasets with relatively lower entropy values (e.g., NASA, BPIC 2012 A, Helpdesk, BPIC 2020), simpler models such as Decision Trees and Random Forests achieved competitive performance compared to deep learning models. By contrast, in datasets exhibiting higher entropy values (e.g., Sepsis, Filtered Hospital Logs), the DAW-Transformer consistently outperformed classical methods, highlighting its ability to capture complex dependencies in unstructured processes.

At the lower end of the entropy spectrum, processes are highly structured and well served by interpretable models. At the higher end, where event logs display significant variability and uncertainty, more expressive deep learning architectures such as the DAW-Transformer become necessary. Table~\ref{tab:process_entropy} summarizes the entropy values of the datasets, illustrating how they align with observed model performance.

\begin{table}[htbp]
    \centering
    \small
    \caption{Conditional and Normalized Entropy of Event Logs}
    \label{tab:process_entropy}
    \begin{tabular}{lcc}
        \toprule
        \textbf{Event Log} & \textbf{Conditional Entropy} & \textbf{Normalized Entropy} \\
        \midrule
        Sepsis & 2.0084 & 0.5021 \\
        Filtered\_Hospital\_Logs & 1.8792 & 0.3132 \\
        NASA & 1.5794 & 0.2843 \\
        BPIC\_2012\_A & 0.7771 & 0.2339 \\
        Help desk & 0.8839 & 0.2322 \\
        BPIC\_2020\_Prepaid Travel Cost & 0.9335 & 0.1922 \\
        \bottomrule
    \end{tabular}

\end{table}

The proposed DAW-Transformer was evaluated against a range of machine learning and deep learning baselines, with the comparative results presented in~\Cref{table:accuracy_comparison_entropy}. Summarizes these evaluation results, highlighting how different models perform across datasets with varying levels of complexity. Among our datasets, Sepsis (0.50) and Filtered Hospital Logs (0.31) exhibited relatively higher entropy values, making them more challenging for interpretable models. In both cases, the DAW-Transformer consistently outperformed the baselines. For the Sepsis dataset, it achieved an accuracy of 64.88\%, while Random Forest and Decision Tree reached 47.93\% and 50.30\%, respectively. Similarly, in the Filtered Hospital Logs, the DAW-Transformer achieved 74.63\%, outperforming Random Forest (65.80\%) and Decision Tree (67.32\%). These differences highlight the advantage of advanced architectures in capturing long-range dependencies in high-entropy processes.

In contrast, performance differences across models were minimal in low-entropy datasets. For example, in the BPIC\_2020\_Prepaid Travel Cost dataset (normalized entropy: 0.19), accuracies were nearly identical: 93.93\% for Decision Tree, 93.76\% for Random Forest, and 94.35\% for the DAW-Transformer. A similar trend was observed in the Helpdesk dataset (normalized entropy: 0.23), where the DAW-Transformer achieved 79.48\%, only slightly higher than XGBoost (79.19\%) and Decision Trees (77.43\%). These results suggest that in highly structured, low-entropy processes, simpler models are sufficient, as they provide competitive accuracy without the computational overhead of deep learning architectures.

\begin{figure*}[htbp]
    \centering
 
    \hspace*{0pt}
    \vspace{10pt}
    \includegraphics[width=1\textwidth]{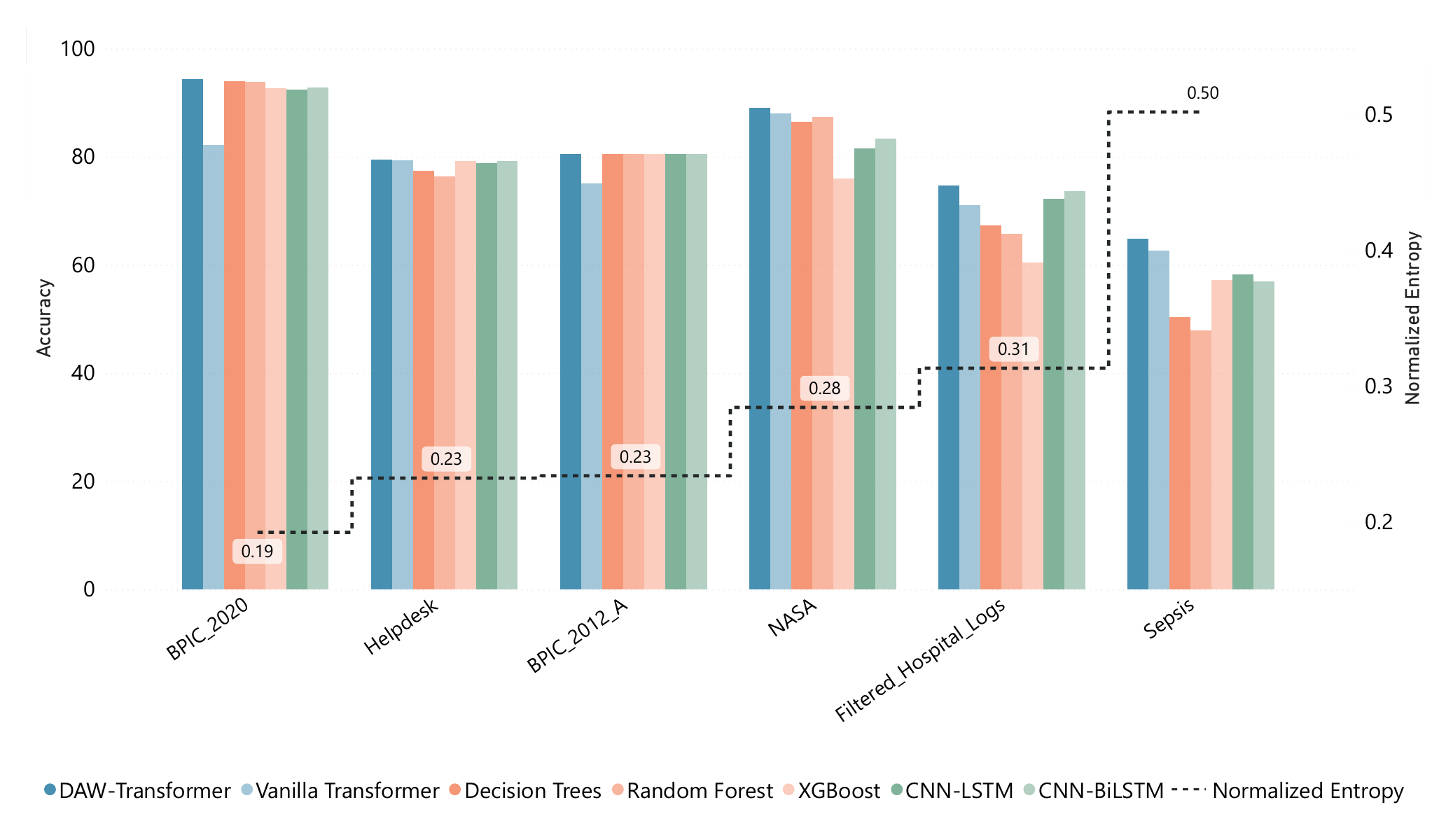}
    \caption{\centering
         Model accuracy across event logs: Decision Tree, Random Forest, and DAW-Transformer perform similarly in low-entropy logs, while DAW-Transformer shows clear advantages in high-entropy logs.}
    \label{fig:results:charts}
\end{figure*}

Overall, the results demonstrate a consistent pattern: the DAW-Transformer yields substantial gains in high-entropy event logs, with improvements of up to 29\% compared to classical models, while in low-entropy datasets the performance of simpler methods converges closely with that of advanced models. Figure~\ref{fig:results:charts} visualizes the relationship between event logs entropy and model performance. The results show that in datasets with relatively lower normalized entropy values, simpler interpretable models such as Decision Trees perform competitively with advanced architectures. However, as entropy increases and processes become more variable and less predictable, the DAW-Transformer achieves substantially higher accuracy than the interpretable baselines.

\begin{landscape}
\begin{table}[htbp]
\centering
\caption{Performance Comparison across Models and Event Logs (in \%)}
\label{table:accuracy_comparison_entropy}

\setlength{\tabcolsep}{5pt} 
\renewcommand{\arraystretch}{1.2} 

\scalebox{0.65}{%
\begin{tabular}{llcccccccc}
\toprule
\textbf{Event Logs} & \textbf{Metric} & \textbf{\shortstack[c]{Normalized\\Entropy}} & \textbf{Decision Trees} & \textbf{Random Forest} & \textbf{XGBoost} & \textbf{CNN-LSTM} & \textbf{CNN-BiLSTM} & \textbf{\shortstack[c]{Vanilla\\Transformer}} & \textbf{DAW-Transformer} \\
\midrule

\multirow{4}{*}{Sepsis}
 & Accuracy   & 0.5021 & 50.30 & 47.93 & 57.16 & 58.18 & 57.48 & 62.58 & \textbf{64.88} \\
 & F1-score   &        & 50.47 & 47.25 & 55.94 & 55.16 & 54.96 & 61.84 & 64.40 \\
 & Recall     &        & 50.30 & 47.93 & 57.16 & 58.18 & 57.48 & 62.58 & 64.88 \\
 & Precision  &        & 50.74 & 47.54 & 59.75 & 56.73 & 55.26 & 63.19 & 65.31 \\

\midrule
\multirow{4}{*}{Filtered\_Hospital Logs}
 & Accuracy   & 0.3132 & 67.32 & 65.80 & 60.40 & 72.26 & 73.66 & 71.08 & \textbf{74.63} \\
 & F1-score   &        & 66.56 & 65.08 & 55.67 & 68.36 & 71.71 & 70.02 & 73.99 \\
 & Recall     &        & 67.32 & 65.80 & 60.40 & 72.26 & 73.66 & 71.08 & 74.63 \\
 & Precision  &        & 65.95 & 64.69 & 59.25 & 67.41 & 72.21 & 71.17 & 74.61 \\

\midrule
\multirow{4}{*}{NASA}
 & Accuracy   & 0.2843 & 86.46 & 87.37 & 76.00 & 81.55 & 83.30 & 88.04 & \textbf{89.02} \\
 & F1-score   &        & 86.66 & 87.47 & 75.51 & 80.08 & 82.77 & 88.22 & 89.25 \\
 & Recall     &        & 86.46 & 87.37 & 76.00 & 81.55 & 83.30 & 88.04 & 89.02 \\
 & Precision  &        & 89.13 & 88.97 & 78.68 & 81.22 & 84.91 & 90.86 & 92.27 \\

\midrule
\multirow{4}{*}{BPIC\_2012\_A}
 & Accuracy   & 0.2339 & 80.46 & 80.46 & 80.47 & 80.47 & 80.47 & 75.10 & \textbf{80.49} \\
 & F1-score   &        & 75.25 & 75.25 & 75.25 & 75.25 & 75.25 & 69.50 & 72.25 \\
 & Recall     &        & 80.47 & 80.47 & 80.47 & 80.47 & 80.47 & 75.10 & 80.47 \\
 & Precision  &        & 72.24 & 72.24 & 72.24 & 72.24 & 72.24 & 67.66 & 72.24 \\

\midrule
\multirow{4}{*}{Helpdesk}
 & Accuracy   & 0.2322 & 77.43 & 76.39 & 79.19 & 78.86 & 79.16 & 79.28 & \textbf{79.48} \\
 & F1-score   &        & 73.61 & 74.21 & 74.40 & 73.50 & 73.85 & 75.38 & 75.14 \\
 & Recall     &        & 77.43 & 76.39 & 79.19 & 78.86 & 79.16 & 79.28 & 79.48 \\
 & Precision  &        & 73.33 & 73.26 & 72.03 & 69.13 & 69.66 & 81.53 & 81.18 \\

\midrule
\multirow{4}{*}{\shortstack[l]{BPIC\_2020\_\\Prepaid Travel Cost}}
 & Accuracy   & 0.1922 & 93.93 & 93.76 & 92.65 & 92.39 & 92.83 & 82.23 & \textbf{94.35} \\
 & F1-score   &        & 92.86 & 92.69 & 91.35 & 90.93 & 91.36 & 79.71 & 93.11 \\
 & Recall     &        & 93.93 & 93.76 & 92.65 & 92.39 & 92.83 & 82.23 & 94.35 \\
 & Precision  &        & 93.19 & 92.98 & 91.74 & 90.10 & 90.54 & 82.21 & 93.44 \\

\bottomrule
\end{tabular}%
}
\end{table}
\end{landscape}

\section{Discussion}

The results confirm that process entropy is a reliable indicator for selecting predictive models in business process management (BPM). By aligning model complexity with dataset entropy, the proposed framework achieves a balance between predictive accuracy and interpretability. In structured, low-entropy settings, models like Decision Trees provide transparent, rule-based logic that domain experts can readily validate. This interpretability enables users to trust predictions, justify decisions, and comply with regulatory requirements. By contrast, although the DAW-Transformer achieves superior accuracy in high-entropy scenarios, its internal mechanisms are less transparent. The entropy-driven framework thus ensures interpretability is preserved whenever possible, while complexity is introduced only when strictly necessary.

\begin{figure*}[ht!]
    \centering
    \begin{subfigure}[b]{0.46\textwidth}
        \centering
        \includegraphics[width=\linewidth]{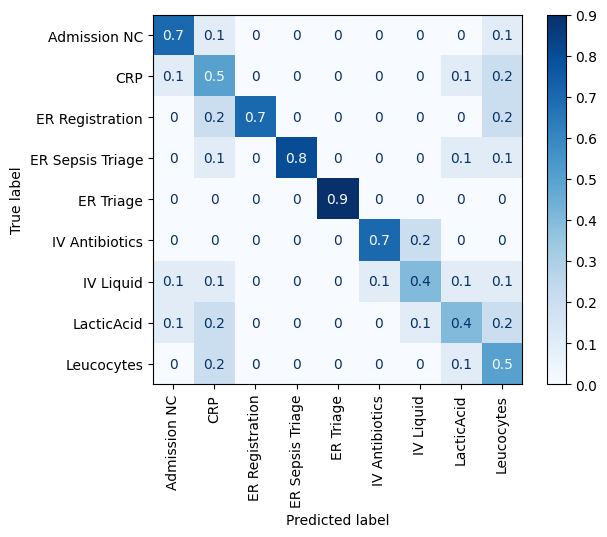}
        \caption{Random Forest}
        \label{fig:second_figure}
    \end{subfigure}
    \hfill
    \begin{subfigure}[b]{0.46\textwidth}
        \centering
        \includegraphics[width=\linewidth]{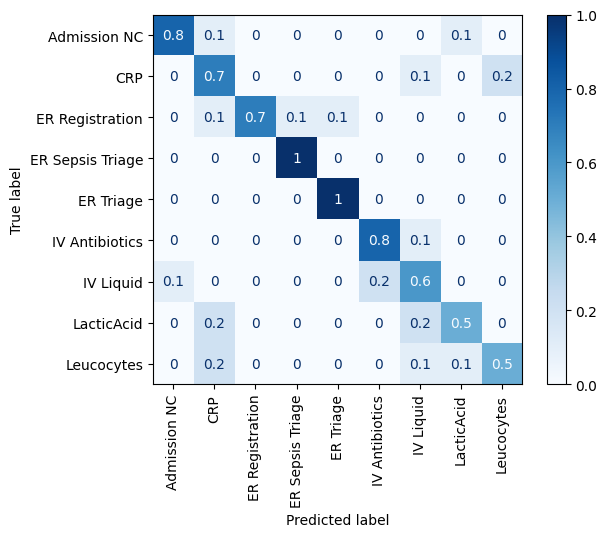}
        \caption{DAW-Transformer}
        \label{fig:confusion_matrix_first}
    \end{subfigure}
    \vspace{10pt} 
    \caption{\centering
        The Sepsis confusion matrix demonstrates improved performance
    }
    \label{fig:side_by_side_figures}
\end{figure*}

Based on results, in high-entropy event logs (Sepsis), the DAW-Transformer yielded a 29\% accuracy improvement over simpler models. Confusion matrix analysis (Figure~\ref{fig:side_by_side_figures}) further illustrates this effect: the DAW-Transformer produced a denser diagonal compared to Random Forest. These findings underscore the necessity of deep architectures in highly variable, unpredictable processes.

\begin{figure*}[htbp]
    \centering
 
    \hspace*{0pt}
    \vspace{10pt}
    \includegraphics[width=1\textwidth]{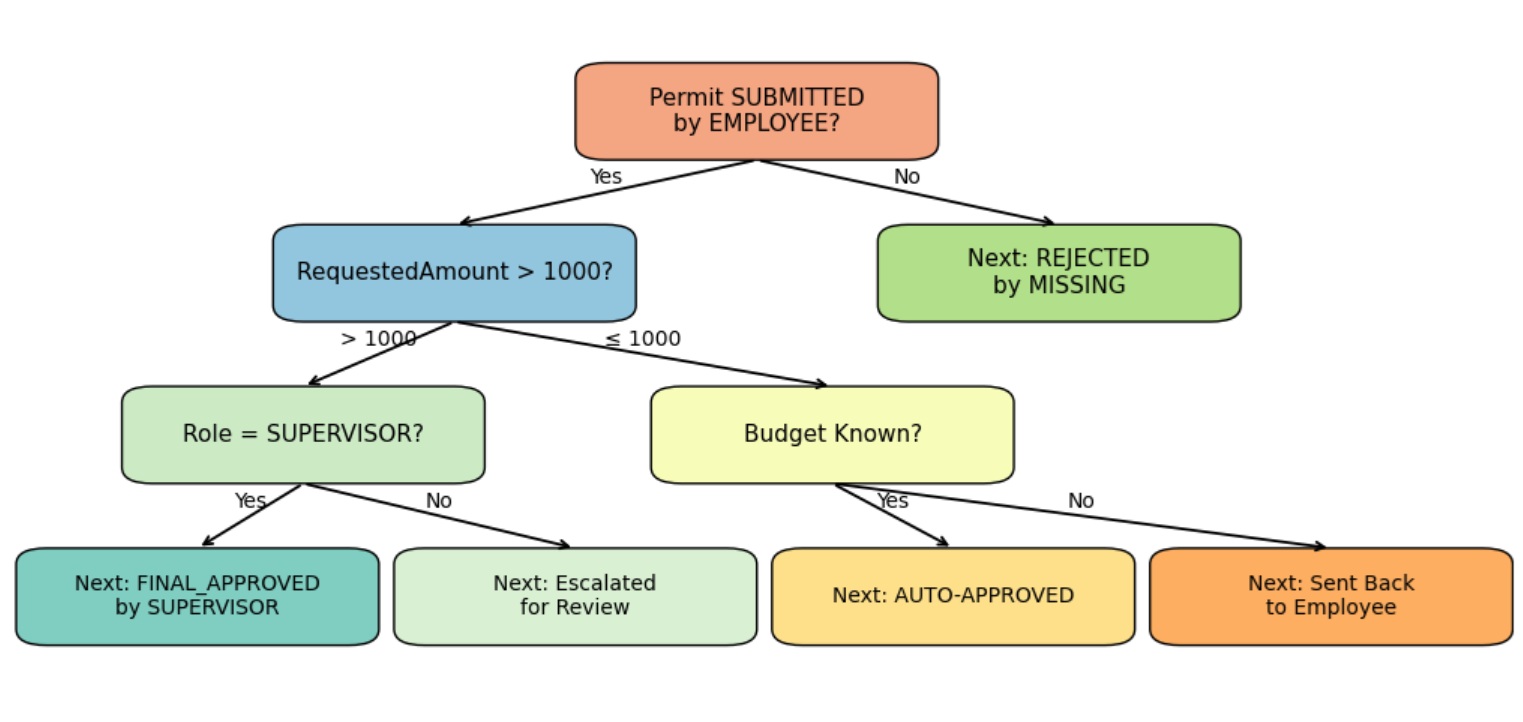}
    \caption{\centering
         3-level decision tree for BPIC\_2020\_Prepaid Travel Cost}
    \label{fig:Decision_Trees}
\end{figure*}

By contrast, in the BPIC\_2020\_Prepaid Travel Cost dataset (normalized entropy = 0.19), a Decision Tree achieved 93.93\% accuracy, which was nearly identical to the DAW-Transformer’s 94.35\%. In this case, the Decision Tree produced simple decision rules. Figure~\ref{fig:Decision_Trees} illustrates these rules in a simplified 3-level Decision Tree for the BPIC 2020 Prepaid Travel Cost dataset. The figure highlights how interpretable, rule-based structures can transparently represent next activity predictions, making them easily understandable for stakeholders.

These rules are transparent and easy to communicate to stakeholders, enabling auditors and managers to directly validate predictions against organizational policies. Similarly, in Helpdesk, all models achieved similarly high accuracy. In such structured settings, simpler approaches like Decision Trees or Random Forests are preferable, as they deliver transparent, rule-based predictions that domain experts can readily validate. Such rule-based outputs demonstrate how Decision Trees can complement DAW-Transformer in practice:

\begin{itemize}
    \item For high-entropy datasets, DAW-Transformer captures complex, long-range dependencies that interpretable models cannot.

    \item For low-entropy datasets, Decision Trees provide competitive accuracy while offering clear, case-level decision rules that enhance stakeholder trust and usability.

\end{itemize}

By presenting both types of results, the entropy-driven framework not only optimizes predictive accuracy but also ensures that interpretability is preserved whenever possible.

Beyond accuracy, we assessed precision, recall, and F1-scores to provide a more complete evaluation (Table~\ref{table:accuracy_comparison_entropy}). These metrics also confirmed the entropy-driven selection model. On high-entropy datasets, the DAW-Transformer consistently maintained balanced precision and recall, achieving the strongest F1-scores (e.g., 64.40\% on Sepsis, 73.99\% on Filtered Hospital Logs). Simpler models, in contrast, exhibited weaker recall, reflecting frequent misclassification of valid activities. In low-entropy datasets, however, differences across models were negligible, with F1-scores varying by less than 2\%.

Taken together, the analysis of precision, recall, and F1-scores reinforces the entropy-driven model selection framework. In structured, low-entropy processes, simpler models provide competitive accuracy and balanced classification performance. However, in unstructured, high-entropy settings, advanced models such as the DAW-Transformer are necessary to achieve both higher accuracy and improved reliability across all evaluation metrics.

The key implication of these findings is that entropy can guide adaptive model selection: employ interpretable models for structured processes, and apply more complex architectures only when required by process complexity. This avoids both underfitting in high-entropy scenarios and unnecessary over-engineering in low-entropy ones.

\section{Conclusions}

This paper introduced an entropy-driven model selection framework for next-activity prediction in business process management (BPM). It proposed the DAW-Transformer, a novel deep learning architecture that leverages dynamic windows and multi-head attention to capture complex dependencies in event logs. Together, these contributions address the trade-off between predictive performance and interpretability by aligning model complexity with the inherent uncertainty of processes.

Experiments on six publicly available event logs confirmed the effectiveness of this dual contribution. For high-entropy datasets such as Sepsis, the DAW-Transformer significantly outperformed both classical machine learning methods and state-of-the-art deep learning baselines. For low-entropy datasets such as BPIC\_2020\_Prepaid Travel Cost, simpler models like decision trees delivered comparable accuracy while providing greater transparency and lower computational costs. These results demonstrate that entropy is a practical criterion for guiding adaptive model selection, ensuring that advanced architectures are employed only where necessary.

The proposed framework offers an explainable and practical approach for guiding model selection in predictive business process monitoring. At present, applying the method benefits from human insight, which reflects on choosing the models. Looking ahead, future work can build on this foundation by developing more automated mechanisms that preserve explainability while reducing the need for human intervention, thereby making the framework even more scalable and accessible.

\section{Acknowledgment}
We acknowledge the support of the Natural Sciences and Engineering Research Council of Canada (NSERC) Discovery Grant No. RGPIN-2023-05408.

\bibliographystyle{elsarticle-num} 
\bibliography{ref.bib}

\begin{thebibliography}{10}
\expandafter\ifx\csname url\endcsname\relax
  \def\url#1{\texttt{#1}}\fi
\expandafter\ifx\csname urlprefix\endcsname\relax\def\urlprefix{URL }\fi
\expandafter\ifx\csname href\endcsname\relax
  \def\href#1#2{#2} \def\path#1{#1}\fi

\bibitem{burattin2015process}
A.~Burattin, Process mining techniques in business environments, Lecture Notes
  in Business Information Processing 207~(1) (2015) 220.

\bibitem{turner2012process}
C.~J. Turner, A.~Tiwari, R.~Olaiya, Y.~Xu, Process mining: from theory to
  practice, Business process management journal 18~(3) (2012) 493--512.

\bibitem{revina2023approach}
A.~Revina, {\"U}.~Aksu, An approach for analyzing business process execution
  complexity based on textual data and event log, Information Systems 114
  (2023) 102184.

\bibitem{dentamaro2023next}
V.~Dentamaro, D.~Impedovo, G.~Pirlo, G.~Semeraro, et~al., Next activity
  prediction and elapsed time prediction on process dataset., in: Ital-IA,
  2023, pp. 605--609.

\bibitem{polato2018time}
M.~Polato, A.~Sperduti, A.~Burattin, M.~d. Leoni, Time and activity sequence
  prediction of business process instances, Computing 100 (2018) 1005--1031.

\bibitem{sun2024next}
X.~Sun, S.~Yang, Y.~Ying, D.~Yu, Next activity prediction of ongoing business
  processes based on deep learning, Expert Systems 41~(5) (2024) e13421.

\bibitem{pufahl2025resource}
L.~Pufahl, F.~Stiehle, S.~Ihde, M.~Weske, I.~Weber, Resource allocation in
  business process executions—a systematic literature study, Information
  Systems (2025) 102541.

\bibitem{middelhuis2025learning}
J.~Middelhuis, R.~L. Bianco, E.~Sherzer, Z.~Bukhsh, I.~Adan, R.~Dijkman,
  Learning policies for resource allocation in business processes, Information
  Systems 128 (2025) 102492.

\bibitem{bukhsh2021processtransformer}
Z.~A. Bukhsh, A.~Saeed, R.~M. Dijkman, Processtransformer: Predictive business
  process monitoring with transformer network, arXiv preprint arXiv:2104.00721
  (2021).

\bibitem{rivera2022multi}
G.~Rivera~Lazo, R.~{\~N}anculef, Multi-attribute transformers for sequence
  prediction in business process management, in: International Conference on
  Discovery Science, Springer, 2022, pp. 184--194.

\bibitem{mehdiyev2020novel}
N.~Mehdiyev, J.~Evermann, P.~Fettke, A novel business process prediction model
  using a deep learning method, Business \& information systems engineering 62
  (2020) 143--157.

\bibitem{abbasi2025forlaps}
M.~Abbasi, M.~Khadivi, M.~Ahang, P.~Lasserre, Y.~Lucet, H.~Najjaran, Forlaps:
  An innovative data-driven reinforcement learning approach for prescriptive
  process monitoring, arXiv preprint arXiv:2501.10543 (2025).

\bibitem{lecun2015deep}
Y.~LeCun, Y.~Bengio, G.~Hinton, Deep learning, nature 521~(7553) (2015)
  436--444.

\bibitem{wang2023mitfm}
J.~Wang, C.~Lu, B.~Cao, J.~Fan, Mitfm: A multi-view information fusion method
  based on transformer for next activity prediction of business processes, in:
  Proceedings of the 14th Asia-Pacific Symposium on Internetware, 2023, pp.
  281--291.

\bibitem{musa2023prediction}
T.~H.~A. Musa, A.~Bouras, Prediction of next events in business processes: A
  deep learning approach, in: IFIP International Conference on Product
  Lifecycle Management, Springer, 2023, pp. 210--220.

\bibitem{di2019activity}
N.~Di~Mauro, A.~Appice, T.~M. Basile, Activity prediction of business process
  instances with inception cnn models, in: AI* IA 2019--Advances in Artificial
  Intelligence: XVIIIth International Conference of the Italian Association for
  Artificial Intelligence, Rende, Italy, November 19--22, 2019, Proceedings 18,
  Springer, 2019, pp. 348--361.

\bibitem{vaswani2017attention}
A.~Vaswani, Attention is all you need, Advances in Neural Information
  Processing Systems (2017).

\bibitem{kumar2024transparency}
J.~R.~R. Kumar, A.~Kalnawat, A.~M. Pawar, V.~D. Jadhav, P.~Srilatha,
  V.~Khetani, Transparency in algorithmic decision-making: Interpretable models
  for ethical accountability, in: E3S Web of Conferences, Vol. 491, EDP
  Sciences, 2024, p. 02041.

\bibitem{impedovo2023next}
D.~Impedovo, G.~Pirlo, G.~Semeraro, Next activity prediction: An application of
  shallow learning techniques against deep learning over the bpi challenge
  2020, IEEE Access 11~(1) (2023) 117947--117953.

\bibitem{weinzierl2020prescriptive}
S.~Weinzierl, S.~Dunzer, S.~Zilker, M.~Matzner, Prescriptive business process
  monitoring for recommending next best actions, in: International conference
  on business process management, Springer, 2020, pp. 193--209.

\bibitem{alaghbari2022activities}
K.~A. Alaghbari, M.~H.~M. Saad, A.~Hussain, M.~R. Alam, Activities recognition,
  anomaly detection and next activity prediction based on neural networks in
  smart homes, IEEE Access 10~(1) (2022) 28219--28232.

\bibitem{ceci2014completion}
M.~Ceci, P.~F. Lanotte, F.~Fumarola, D.~P. Cavallo, D.~Malerba, Completion time
  and next activity prediction of processes using sequential pattern mining,
  in: Discovery Science: 17th International Conference, DS 2014, Bled,
  Slovenia, October 8-10, 2014. Proceedings 17, Springer, 2014, pp. 49--61.

\bibitem{breiman2001random}
L.~Breiman, Random forests, Machine learning 45~(1) (2001) 5--32.

\bibitem{song2015decision}
Y.-Y. Song, L.~Ying, Decision tree methods: applications for classification and
  prediction, Shanghai archives of psychiatry 27~(2) (2015) 130.

\bibitem{conforti2015recommendation}
R.~Conforti, M.~De~Leoni, M.~La~Rosa, W.~M. Van Der~Aalst, A.~H. Ter~Hofstede,
  A recommendation system for predicting risks across multiple business process
  instances, Decision Support Systems 69 (2015) 1--19.

\bibitem{di2018predictive}
C.~Di~Francescomarino, C.~Ghidini, F.~M. Maggi, F.~Milani, Predictive process
  monitoring methods: Which one suits me best?, in: International conference on
  business process management, Springer, 2018, pp. 462--479.

\bibitem{leitner2013data}
P.~Leitner, J.~Ferner, W.~Hummer, S.~Dustdar, Data-driven and automated
  prediction of service level agreement violations in service compositions,
  Distributed and Parallel Databases 31~(3) (2013) 447--470.

\bibitem{tama2019empirical}
B.~A. Tama, M.~Comuzzi, An empirical comparison of classification techniques
  for next event prediction using business process event logs, Expert Systems
  with Applications 129 (2019) 233--245.

\bibitem{kang2012periodic}
B.~Kang, D.~Kim, S.-H. Kang, Periodic performance prediction for real-time
  business process monitoring, Industrial Management \& Data Systems 112~(1)
  (2012) 4--23.

\bibitem{verenich2016minimizing}
I.~Verenich, M.~Dumas, M.~La~Rosa, F.~M. Maggi, C.~Di~Francescomarino,
  Minimizing overprocessing waste in business processes via predictive activity
  ordering, in: International conference on advanced information systems
  engineering, Springer, 2016, pp. 186--202.

\bibitem{marquez2017run}
A.~E. M{\'a}rquez-Chamorro, M.~Resinas, A.~Ruiz-Cort{\'e}s, M.~Toro, Run-time
  prediction of business process indicators using evolutionary decision rules,
  Expert Systems with Applications 87 (2017) 1--14.

\bibitem{abb2023discussion}
L.~Abb, P.~Pfeiffer, P.~Fettke, J.-R. Rehse, A discussion on generalization in
  next-activity prediction, in: International Conference on Business Process
  Management, Springer, 2023, pp. 18--30.

\bibitem{abbasi2024review}
M.~Abbasi, R.~I. Nishat, C.~Bond, J.~B. Graham-Knight, P.~Lasserre, Y.~Lucet,
  H.~Najjaran, A review of ai and machine learning contribution in business
  process management (process enhancement and process improvement approaches),
  Business Process Management Journal (2024).

\bibitem{weinzierl2020xnap}
S.~Weinzierl, S.~Zilker, J.~Brunk, K.~Revoredo, M.~Matzner, J.~Becker, Xnap:
  making lstm-based next activity predictions explainable by using lrp, in:
  International Conference on Business Process Management, Springer, 2020, pp.
  129--141.

\bibitem{krishna2018lstm}
K.~Krishna, D.~Jain, S.~V. Mehta, S.~Choudhary, An lstm based system for
  prediction of human activities with durations, Proceedings of the ACM on
  Interactive, Mobile, Wearable and Ubiquitous Technologies 1~(4) (2018) 1--31.

\bibitem{irbaz2024predicting}
M.~S. Irbaz, L.~N. Lota, F.~A. Sakib, Predicting user-specific future
  activities using lstm-based multi-label classification, in: Human Activity
  and Behavior Analysis, CRC Press, 2024, pp. 301--310.

\bibitem{tax2017predictive}
N.~Tax, I.~Verenich, M.~La~Rosa, M.~Dumas, Predictive business process
  monitoring with lstm neural networks, in: International conference on
  advanced information systems engineering, Springer, 2017, pp. 477--492.

\bibitem{rama2021deep}
E.~Rama-Maneiro, J.~C. Vidal, M.~Lama, Deep learning for predictive business
  process monitoring: Review and benchmark, IEEE Transactions on Services
  Computing 16~(1) (2021) 739--756.

\bibitem{aversano2023data}
L.~Aversano, M.~L. Bernardi, M.~Cimitile, M.~Iammarino, C.~Verdone, A
  data-aware explainable deep learning approach for next activity prediction,
  Engineering Applications of Artificial Intelligence 126~(1) (2023) 106758.

\bibitem{wang2023mtlformer}
J.~Wang, J.~Huang, X.~Ma, Z.~Li, Y.~Wang, D.~Yu, Mtlformer: Multi-task learning
  guided transformer network for business process prediction, IEEE Access
  (2023).

\bibitem{raschka2018model}
S.~Raschka, Model evaluation, model selection, and algorithm selection in
  machine learning, arXiv preprint arXiv:1811.12808 (2018).

\bibitem{domingos2012few}
P.~Domingos, A few useful things to know about machine learning, Communications
  of the ACM 55~(10) (2012) 78--87.

\bibitem{kim2024explaining}
S.~Kim, M.~Comuzzi, C.~Di~Francescomarino, Explaining the impact of design
  choices on model quality in predictive process monitoring, Journal of
  Intelligent Information Systems (2024) 1--26.

\bibitem{drodt2023predictive}
C.~Drodt, S.~Weinzierl, M.~Matzner, P.~Delfmann, Predictive recommining:
  Learning relations between event log characteristics and machine learning
  approaches for supporting predictive process monitoring, in: International
  Conference on Advanced Information Systems Engineering, Springer, 2023, pp.
  69--76.

\bibitem{jung2008measuring}
J.-Y. Jung, Measuring entropy in business process models, in: 2008 3rd
  International Conference on Innovative Computing Information and Control,
  IEEE, 2008, pp. 246--246.

\bibitem{jung2011entropy}
J.-Y. Jung, C.-H. Chin, J.~Cardoso, An entropy-based uncertainty measure of
  process models, Information Processing Letters 111~(3) (2011) 135--141.

\bibitem{rahane2020measures}
A.~A. Rahane, A.~Subramanian, Measures of complexity for large scale image
  datasets, in: 2020 international conference on artificial intelligence in
  information and communication (ICAIIC), IEEE, 2020, pp. 282--287.

\bibitem{huang2021deep}
C.~Huang, A.~Deep, S.~Zhou, D.~Veeramani, A deep learning approach for
  predicting critical events using event logs, Quality and Reliability
  Engineering International 37~(5) (2021) 2214--2234.

\bibitem{bolt2017finding}
A.~Bolt, W.~M. van~der Aalst, M.~De~Leoni, Finding process variants in event
  logs: (short paper), in: On the Move to Meaningful Internet Systems. OTM 2017
  Conferences: Confederated International Conferences: CoopIS, C\&TC, and
  ODBASE 2017, Rhodes, Greece, October 23-27, 2017, Proceedings, Part I,
  Springer, 2017, pp. 45--52.

\bibitem{wilcox1967indices}
A.~R. Wilcox, Indices of qualitative variation., Tech. rep., Oak Ridge National
  Lab.(ORNL), Oak Ridge, TN (United States) (1967).

\bibitem{https://doi.org/10.4121/uuid:915d2bfb-7e84-49ad-a286-dc35f063a460}
F.~Mannhardt, \href{https://data.4tu.nl/articles/_/12707639/1}{Sepsis cases -
  event log} (2016).
\newblock \href
  {https://doi.org/10.4121/UUID:915D2BFB-7E84-49AD-A286-DC35F063A460}
  {\path{doi:10.4121/UUID:915D2BFB-7E84-49AD-A286-DC35F063A460}}.
\newline\urlprefix\url{https://data.4tu.nl/articles/_/12707639/1}

\bibitem{vandongen2011hospital}
B.~van Dongen,
  \href{https://doi.org/10.4121/uuid:d9769f3d-0ab0-4fb8-803b-0d1120ffcf54}{Real-life
  event logs - hospital log}, dataset (2011).
\newblock \href
  {https://doi.org/10.4121/uuid:d9769f3d-0ab0-4fb8-803b-0d1120ffcf54}
  {\path{doi:10.4121/uuid:d9769f3d-0ab0-4fb8-803b-0d1120ffcf54}}.
\newline\urlprefix\url{https://doi.org/10.4121/uuid:d9769f3d-0ab0-4fb8-803b-0d1120ffcf54}

\bibitem{leemans2017nasa}
M.~Leemans,
  \href{https://doi.org/10.4121/uuid:60383406-ffcd-441f-aa5e-4ec763426b76}{Nasa
  crew exploration vehicle (cev) software event log}, dataset (2017).
\newblock \href
  {https://doi.org/10.4121/uuid:60383406-ffcd-441f-aa5e-4ec763426b76}
  {\path{doi:10.4121/uuid:60383406-ffcd-441f-aa5e-4ec763426b76}}.
\newline\urlprefix\url{https://doi.org/10.4121/uuid:60383406-ffcd-441f-aa5e-4ec763426b76}

\bibitem{vandongen2012BPI2012}
B.~van Dongen,
  \href{https://doi.org/10.4121/uuid:3926db30-f712-4394-aebc-75976070e91f}{Bpi
  challenge 2012}, dataset (2012).
\newblock \href
  {https://doi.org/10.4121/uuid:3926db30-f712-4394-aebc-75976070e91f}
  {\path{doi:10.4121/uuid:3926db30-f712-4394-aebc-75976070e91f}}.
\newline\urlprefix\url{https://doi.org/10.4121/uuid:3926db30-f712-4394-aebc-75976070e91f}

\bibitem{https://doi.org/10.17632/39bp3vv62t.1}
I.~Verenich, \href{https://doi.org/10.17632/39bp3vv62t.1}{Helpdesk}, available
  at: \url{https://doi.org/10.17632/39bp3vv62t.1} (2016).
\newblock \href {https://doi.org/10.17632/39bp3vv62t.1}
  {\path{doi:10.17632/39bp3vv62t.1}}.
\newline\urlprefix\url{https://doi.org/10.17632/39bp3vv62t.1}

\bibitem{https://doi.org/10.4121/uuid:5d2fe5e1-f91f-4a3b-ad9b-9e4126870165}
B.~van Dongen, \href{https://data.4tu.nl/articles/_/12696722/1}{Bpi challenge
  2020: Prepaid travel costs} (2020).
\newblock \href
  {https://doi.org/10.4121/UUID:5D2FE5E1-F91F-4A3B-AD9B-9E4126870165}
  {\path{doi:10.4121/UUID:5D2FE5E1-F91F-4A3B-AD9B-9E4126870165}}.
\newline\urlprefix\url{https://data.4tu.nl/articles/_/12696722/1}

\bibitem{evermann2016deep}
J.~Evermann, J.-R. Rehse, P.~Fettke, A deep learning approach for predicting
  process behaviour at runtime, in: International Conference on Business
  Process Management, Springer, 2016, pp. 327--338.

\bibitem{khan2021deepprocess}
A.~Khan, H.~Le, K.~Do, T.~Tran, A.~Ghose, H.~Dam, R.~Sindhgatta, Deepprocess:
  supporting business process execution using a mann-based recommender system,
  in: International Conference on Service-Oriented Computing, Springer, 2021,
  pp. 19--33.

\bibitem{foody2023challenges}
G.~M. Foody, Challenges in the real world use of classification accuracy
  metrics: From recall and precision to the matthews correlation coefficient,
  Plos one 18~(10) (2023) e0291908.

\end{thebibliography}

\end{document}